\documentclass{article}

\usepackage{microtype}
\usepackage{graphicx}
\usepackage{subcaption}
\usepackage{booktabs} % for professional tables
\usepackage{multirow}
\usepackage{colortbl}
\usepackage[table]{xcolor}
\usepackage{hyperref}

\usepackage[accepted]{icml2026}

\usepackage{amsmath}
\usepackage{amssymb}
\usepackage{mathtools}
\usepackage{amsthm}

\usepackage[capitalize,noabbrev]{cleveref}

\theoremstyle{plain}

\theoremstyle{definition}

\theoremstyle{remark}

\usepackage[textsize=tiny]{todonotes}

\icmltitlerunning{Revitalizing the Beginning: Avoiding Storage Dependency for Model Merging in Continual Learning}

\begin{document}

\twocolumn[
  \icmltitle{Revitalizing the Beginning: Avoiding Storage Dependency \\ for Model Merging in Continual Learning}



  \icmlsetsymbol{equal}{*}

 \begin{icmlauthorlist}
  \icmlauthor{Xi Wang}{xdu}
  \icmlauthor{Cheng Deng}{xdu}
\end{icmlauthorlist}

\icmlaffiliation{xdu}{School of Electronic Engineering, Xidian University, Xi'an, Shaanxi, China}

  \icmlcorrespondingauthor{Cheng Deng}

  \icmlkeywords{Continual Learning}

  \vskip 0.3in
]



\printAffiliationsAndNotice{}  % no special notice (required even if empty)

\begin{abstract}
Model merging provides a compelling paradigm for integrating specialized expertise into a unified multi-task model, a goal that aligns naturally with the sequential knowledge acquisition in continual learning (CL). However, the requirement for preserving diverse forms of previous knowledge conflicts with the storage limitations inherent to CL. In this paper, we systematically analyze existing model merging methods under the constraints of CL. We find that current methods prioritize global alignment, which often leads to the accumulation and amplification of task-specific errors within the continuous data stream; and the vanishing gradients at the onset of subsequent tasks frequently cause optimization to stagnate. These leave the merged model in a suboptimal state at the beginning of the next training phase. To address these challenges, we propose Trajectory Regularized Merging (TRM), a framework that reformulates the merging phase as an optimization process within an augmented  trajectory subspace. Our framework integrates three synergistic objectives including task alignment, prediction consistency, and gradient responsiveness to concurrently preserve merged model's historical stability and re-activate optimization dynamics. Extensive experimental results demonstrate that our method achieves state-of-the-art performance across multiple benchmarks.
\end{abstract}

\section{Introduction}
The rapid evolution of data in the real world presents ongoing challenges for deep learning models. Even for pre-trained models (PTMs), training on dynamic data streams often biases them toward new task data, leading to progressive degradation in performance on previously learned knowledge, a phenomenon known as catastrophic forgetting \cite{zhou2024class}. Continual Learning (CL) seeks to mitigate this by navigating the stability-plasticity dilemma, ensuring the retention of historical expertise while facilitating the acquisition of new capabilities. Recent advancements in CL for PTMs have predominantly leveraged Parameter-Efficient finetuning (PEFT), employing modular components such as prompts \cite{wang2022learning, smith2023coda} or adapters \cite{huang2024class} to isolate task-specific updates. More recently, model merging \cite{DBLP:conf/iclr/IlharcoRWSHF23, wortsman2022model} has emerged as a training-free paradigm for model adapting, task-specific models are integrated into a unified multi-task model by selecting or interpolating task vectors, which capture the parameter differences between finetuned and initial model across tasks. 

However, the most existing merging methods \cite{wortsman2022model,jang2024model,DBLP:conf/iclr/IlharcoRWSHF23,yadav2023ties} require access to all previous task vectors and the initial model during the merging phase. This operational requirement is functionally equivalent to storing the entire ensemble of prior experts, which fundamentally contradicts the memory constraints of CL that prohibit the retention of historical data or models to ensure scalability and privacy. Although methods specifically designed for CL, such as MagMax \cite{marczak2024magmax} and BECAME \cite{li2025became}, attempt to mitigate this storage overhead, they still retain accumulated task vectors or Fisher information matrices, which still presents the aforementioned problems in terms of memory. Our experiments reveal that under a strictly CL constraint, where only the current expert and the immediately merged model are available, existing methods suffer from a catastrophic performance degradation of up to $6\%$. (See the supplementary material for detailed results.) These findings prompted us to further investigate whether there are more effective model merging methods for continual learning under strict data storage constraints. 

In alignment with the insights from \cite{DBLP:conf/cvpr/DziadzioURPAAB25}, performance variation of merged model is primarily influenced by the initial model in each training task. In this paper, we first design a series of experiments to examine the stability and plasticity of merged models. We reveal that existing merging methods predominantly optimize for task-agnostic global alignment while neglecting task-specific local optimality. This oversight triggers a progressive representational drift, where localized errors are compounded throughout sequential training, ultimately destabilizing the model's retention of prior knowledge. Furthermore, the merged models frequently exhibit optimization stagnation during the early phases of new task adaptation, indicating a loss of parameter plasticity. To jointly address these challenges, we propose Trajectory Regularized Merging (TRM). Our framework reformulates the merging phase as a guided optimization problem within an augmented trajectory subspace, which is spanned solely by the task vectors of the models before and after training. And we introduce a multi objective supervisory signal composed of three synergistic constraints, task alignment for localized precision, prediction consistency for structural stability, and gradient responsiveness for kinetic re-activation. By navigating this regularized trajectory, TRM identifies an optimal merging point that harmonizes historical stability with future plasticity. This enables robust knowledge integration across dynamic data streams without any reliance on any stored models or data replay.

Our main contributions can be summarized as follows:
\begin{itemize}
    \item We pointed out that existing model merging methods do not fully adhere to the principles of continual learning, and their performance degraded significantly when knowledge storage is disallowed.
    \item We analyzed the cause of this performance gap and reformulated the merging phase as an optimal point search problem within an orthogonally augmented  trajectory subspace.
    \item We proposed an objective composed of three constraints to guide the optimize for the optimal merging point.
    \item Our method achieves state-of-the-art performance across multiple benchmarks.
\end{itemize}

\section{Related Work}
\subsection{Continual Learning}
Continual learning aims to acquire new knowledge from a never-ending data stream continuously \cite{wang2022improving, zhu2021class, zhao2023does, wang2025class}. The primary challenge is learning without catastrophic forgetting: as new data arrives, the model's performance on previously learned tasks should not degrade significantly \cite{li2017learning, CIL0103}. Traditional Regularization based methods \cite{zenke2017continual, aljundi2018memory} penalize changes to important parameters for previous tasks to mitigate forgetting. Conversely, rehearsal based methods \cite{rebuffi2017icarl, lopez2017gradient, buzzega2020dark} maintain a small episodic memory of past data or employ generative models to replay synthetic samples. Recently, with the widespread adoption of pre-trained models, many continual learning methods have been developed as extensions of parameter-efficient finetuning (PEFT) methods. Prompt-based methods \cite{wang2022learning, wang2022dualprompt, smith2023coda, qiao2024prompt, le2024mixture} have demonstrated the effectiveness of migrating pre-trained models into continuous data streams, adapter-based methods \cite{huang2024class, gao2024beyond, tan2024semantically, yu2024boosting, liu2023tail}, achieved high performance by training only a small number of parameters. Additionally, there are methods \cite{khan2023introducing, yu2025language} that consider the knowledge of language modality to aid in modeling learning. 
\subsection{Model Merging}
Model merging has recently gained significant attention as a practical technique for aggregating multiple models by performing linear interpolation in parameter space \cite{xu2024training}. The core idea traces back to ensemble learning methods such as Bagging Predictors \cite{breiman1996bagging}, which improve generalization by averaging the outputs of diverse models. Stochastic Weight Averaging \cite{izmailov2018averaging} aggregates gradients over training to yield wider optima and improved robustness. 
Subsequently, recent methods shift the focus from output space to a weight space combination. numerous methods \cite{wortsman2022model,jang2024model,DBLP:conf/iclr/IlharcoRWSHF23,yadav2023ties} have explored weight combination to achieve model merging, and MagMax \cite{marczak2024magmax} extends model merging to the continual learning, achieving excellent performance through the appropriate storage of model parameters.

\section{Background and Motivation}
\subsection{Problem Formulation}
We consider a supervised continual learning based on pre-trained models, where a pre-trained model (PTM) $f(:,\theta_{init})$ parametrized by $\theta_{init}$ is required to learn a sequence of $\mathcal{T}$ tasks in order. Each task dataset $\mathcal{D}^t$ contains different classes $C^t$ and there is no overlapping between any two different tasks, i.e., $C^i \cap C^j = \emptyset$, $i \neq j$ and $(x,y) \in D^t$ denotes the training sample in task $t$. For each task $t$, training is strictly initialized from the model $f(:,\theta_{t-1})$obtained after task $t-1$, and any previous models or knowledge are permitted to be stored in any form except for $f(:,\theta_{init})$. For simplicity, $\theta_{i}$ will represents the $i^{th}$ model in the following content for convenience. 

Following the conventions in model merging \cite{wortsman2022model}, the task vector $\tau$ is defined as the parameter space displacement from the initial PTM. Specifically, at task $t$, the task vector for the current model $\theta_t$ is $\tau_t = \theta_t - \theta_{init}$. For the model obtained after task $t$, we use superscripts to distinguish between the finetuned $\tilde{\theta_{t}}$ and merged models $\theta_{t}$.

Next, we examine why using the merged model as the initialization for the next training stage has such a substantial impact on overall performance.
\subsection{Analysis}
To uncover the underlying mechanisms of merging failure under strict CL constraints, we establish a controlled experimental protocol using three disjoint tasks $\{\mathcal{T}_0, \mathcal{T}_1, \mathcal{T}_2\}$ derived from ImageNet-R, each comprising 20 non-overlapping classes. We employ $\tilde{\theta_0}$ trained on $\mathcal{T}_0$ as the operational foundation for sequential learning and evaluate three representative merging paradigms: MagMax \cite{marczak2024magmax}, TIES \cite{DBLP:conf/iclr/IlharcoRWSHF23}, and Model Stock \cite{jang2024model}. Next, we design diagnostic experiments grounded in representation stability and optimization plasticity of the merged model to investigate the underlying causes of model failure.
\begin{figure}[h]
    \centering
    \includegraphics[width=0.95\columnwidth]{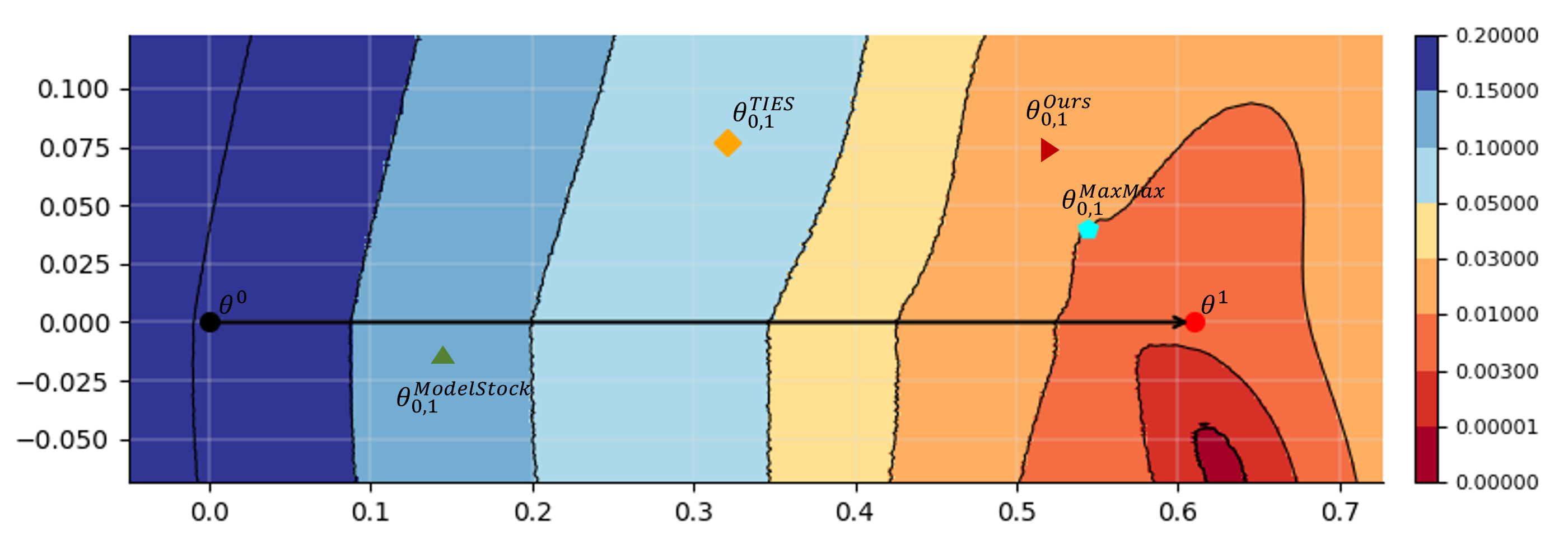}
    \caption{Loss landscape. We visualized the loss landscape along the trajectory from $\tilde{\theta_0}$ to $\tilde{\theta_{1}}$ and projected all merged points onto this surface based on their corresponding loss values.}
    \label{fig2}
\end{figure}
\subsubsection{Suboptimal Local Convergence} 
To visualize the degree of local optimality preservation, we finetune $\tilde{{\theta}_0}$ on $\mathcal{T}_1$ to obtain $\tilde{{\theta}_{1}}$, and get the merged model $\theta_{0,1}^{\zeta }$, where $\zeta \in \{\text{TIES, Model Stock, MagMax}\}$. We project $\theta_{0,1}^{\zeta }$ onto the loss landscape over $\mathcal{T}_1$. As illustrated in Figure \ref{fig2}, a consistent pattern emerges across all evaluated paradigms that the merged models invariably reside in high loss regimes, far removed from the optimal basins of $\mathcal{T}_1$. This indicates that while existing heuristics successfully navigate global parameter space distances, they fundamentally fail to satisfy the local constraints of individual tasks. Formally, we conclude that \textbf{current task-agnostic optimization objectives inadvertently sacrifice task-specific local optimality, leading to the progressive amplification of localized errors across sequential training stages.}
\begin{figure}[h]
    \centering
    \includegraphics[width=0.95\columnwidth]{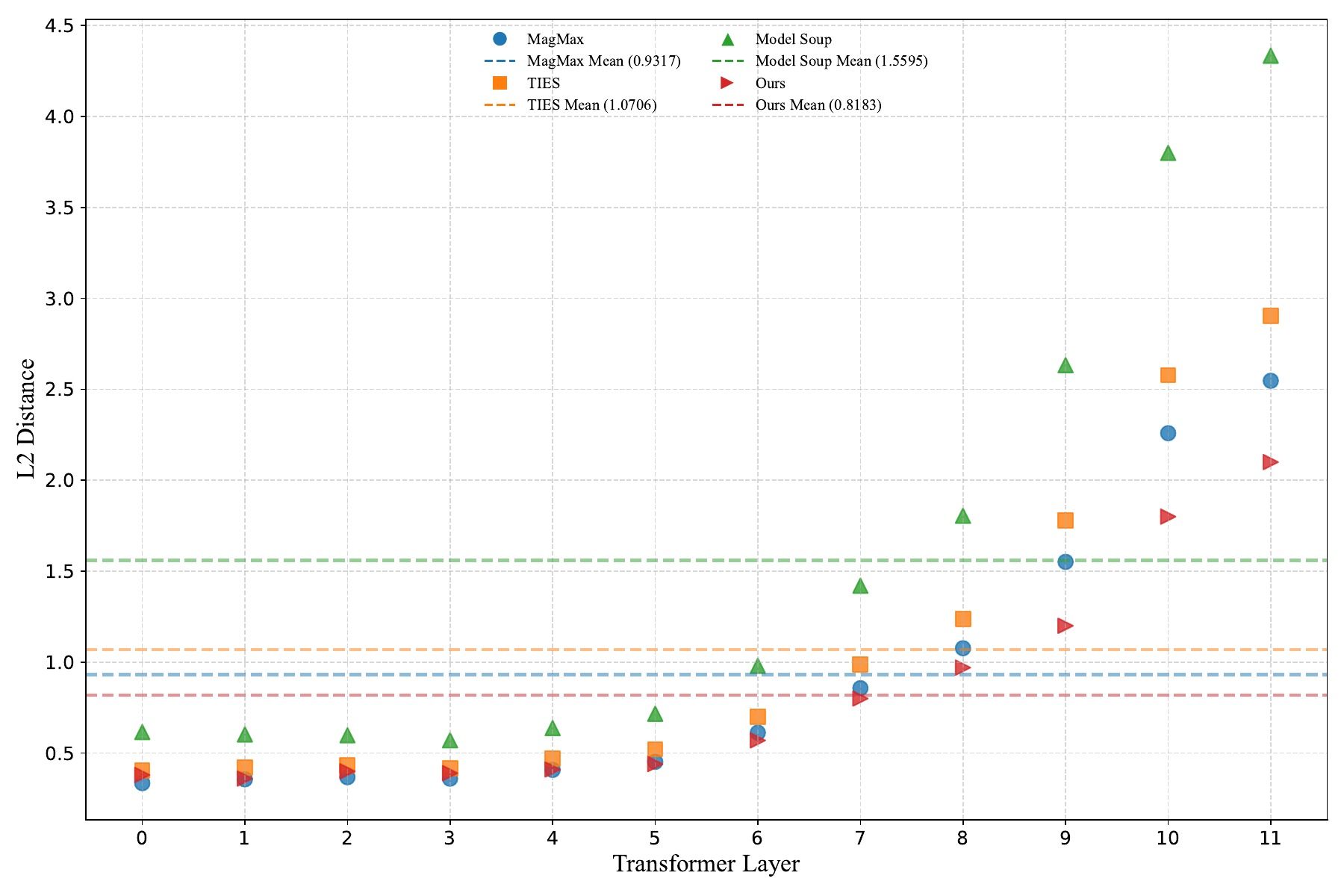}
    \caption{Output drift between $\tilde{\theta_{1}}$ and $\theta_{0,1}^{\zeta }$. We visualize the differences between different layers of the model for the same input.}
    \label{fig3}
\end{figure}
\subsubsection{Disruption of Structural Semantic Representation}
Beyond local optimality, we investigate the internal representational discrepancy between different layers of $\tilde{\theta_{1}}$ and $\theta_{0,1}^{\zeta }$. For any input $x$, the activation deviation at layer $l$ is quantified as the expected $L_2$ distance between their respective hidden representations
\begin{equation}
\Delta_{out} ^l = \frac { 1 } { |\mathcal{D}^1| } \sum _ { j = 1 } ^ { |\mathcal{D}^1| } \left| h^l(x_j; \theta_{0,1}^\zeta) - h^l(x_j; \tilde{\theta_1}) \right| _ 2, \label{eq1}
\end{equation}
where $h^l(\cdot; \theta)$ denotes the hidden states of the $l$-th layer. As shown in Figure \ref{fig3}, the drift is particularly pronounced in deeper layers. Combined with the established observation that lower layers encode general information while higher layers capture task-specific representations \cite{DBLP:conf/iclr/ZhengCQ025}, this finding suggests that vanilla merging methods induces more than a mere numerical perturbation; it substantially disrupts the model’s ability to capture task-specific semantics. We believe that \textbf{well-optimized model parameters exhibit a high degree of structural co-adaptation; linear merging shatters these delicate functional dependencies, triggering catastrophic semantic drift.}
% \begin{equation}
% \Delta_{out} ^{l} = \frac { 1 } { |D^1| } \sum _ { j = 1 } ^ { |D^1| } \left\| {\theta_{0,1}^{\zeta }} ( x _ { j } ) _ { [ 0 ] }^l - {\theta}_{1} ( x _ { j } ) _ { [ 0 ] }^l \right\| _ { 2 },
% \label{eq1}
% \end{equation}
% where $l$ is the corresponding layer, $|D^1|$ is the number of samples of $T^1$ and $[0]$ indicates the class token. The results are shown in Figure \ref{fig3}. It can be observed that existing merging methods alter the model's output under the same input, and this difference becomes increasingly pronounced in deeper layers. We believe that model parameters are highly interdependent after optimization, and the merging phase may introduce abrupt changes in certain parameters. This disruption affects the structured semantic representations encoded in the parameter space, ultimately leading to output drift. Meanwhile, the varying effects of such perturbations across layers support the conclusion previously presented in \cite{DBLP:conf/iclr/ZhengCQ025}, the lower layers of the model encode general information, while the higher layers encode task-specific information.
\subsubsection{Loss of Optimization Plasticity}
\begin{figure}[t]
    \centering
    \includegraphics[width=1.05\columnwidth]{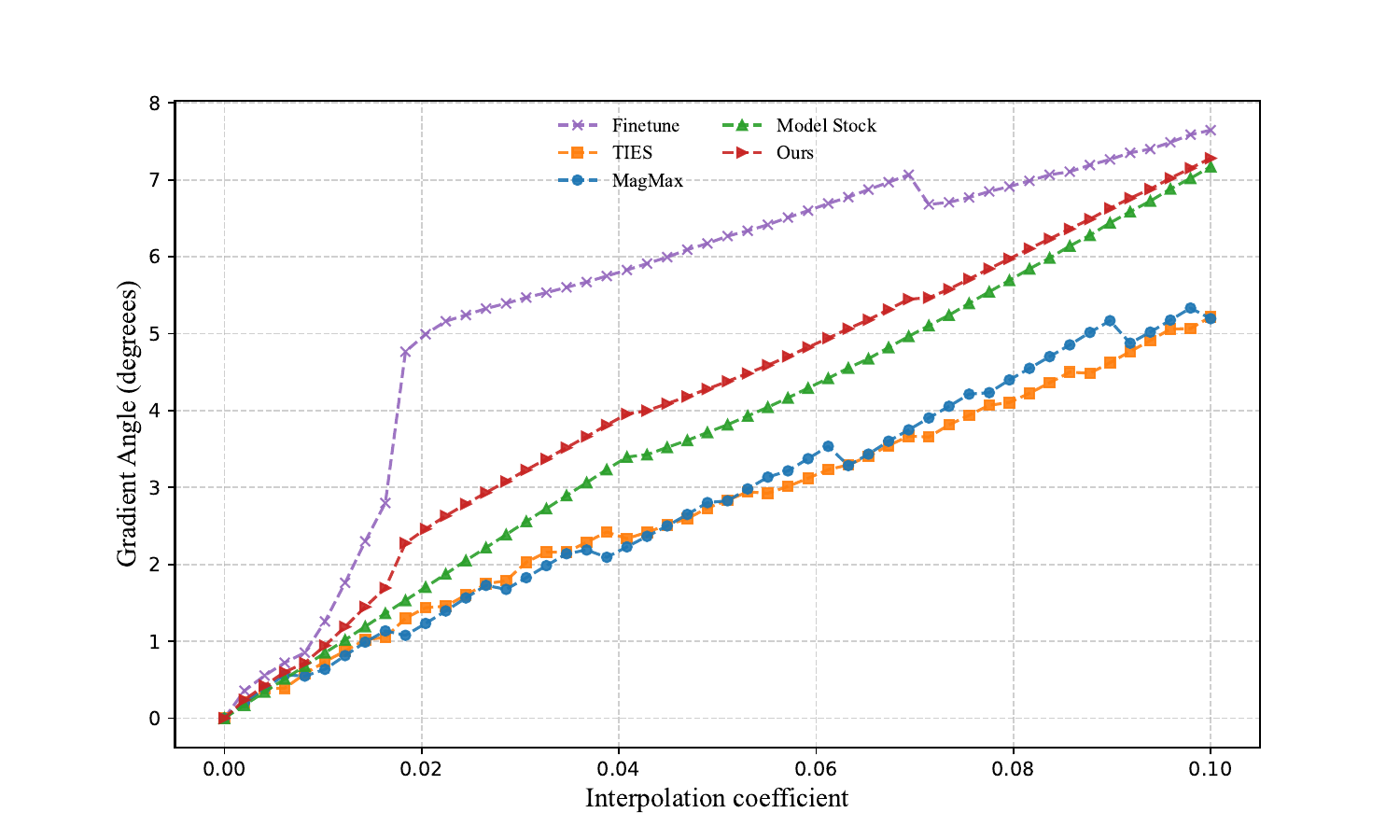}
    \caption{Gradient variations. We visualize the gradient variations between the initial point and neighboring points along the training trajectory.}
    \label{fig4}
\end{figure}

To characterize the optimization dynamics on subsequent tasks, we evaluate the gradient field's sensitivity along the training trajectory of $\mathcal{T}_2$.  We use $\tilde{\theta_{1}}$ and $\theta_{0,1}^{\zeta }$ as the initialization, perform finetuning on $\mathcal{T}_2$ to obtain the corresponding $\tilde{\theta_{2}}$, and then interpolate along the corresponding training trajectory $\tilde{\theta_{1}} \rightarrow \tilde{\theta_{2}}$ and  $\theta_{0,1}^{\zeta } \rightarrow \tilde{\theta_{2}}$ to quantify the gradient angular deviation $\Delta_{\theta}(\delta)$ between the initialization and its neighboring points,
\begin{equation} 
\Delta_{\theta}(\delta) = \arccos \left( \frac{\langle \nabla _ { \theta } \mathcal { L } ( \theta ) , \nabla _ { \theta } \mathcal { L } ( \theta + \delta ) \rangle}{|\nabla _ { \theta } \mathcal { L } ( \theta )|_2 |\nabla _ { \theta } \mathcal { L } ( \theta + \delta )|_2} \right), 
\end{equation}
where $\mathcal { L }$ indicates the loss function and $\delta$ is the perturbation on the training trajectory. As illustrated in Figure \ref{fig4}, the merged model exhibits a strikingly lower angular deviation compared to the directly finetuned baseline. This "directional stiffening" of the gradient suggests that the merged model is trapped in a pathologically flat or saturated region of the loss landscape. In such regimes, the lack of local curvature prevents the optimizer from identifying effective descent directions. We conclude that \textbf{insufficient gradient sensitivity induces a state of kinetic dormancy, which severely erodes the model’s plasticity and its capacity to rapidly adapt to similar, non-stationary data streams.}
Based on these observations, we propose the Trajectory Regularized Merging (TRM) framework, which jointly addresses the aforementioned limitations through three complementary regularization terms.
\section{Method}
We reformulate model merging as a guided optimization problem within an augmented trajectory subspace. The objective is to identify a consolidated parameter state that simultaneously safeguards historical knowledge stability and re-activates optimization plasticity for future tasks.

In the training phase, finetuning starts from $\theta_{t-1}:= merge(\theta_{t-2},\widetilde{\theta_{t-1}})$, where $\widetilde{\theta_{t-1}}$ was obtained by directly finetuning on task $t-1$  and produces model $\widetilde{\theta_{t}}$ under the supervision of the cross-entropy loss $\mathcal { L } _ { c e }$. In the merging phase, we first construct the task vectors $\tau_{t-1}$ and $\tau_{t}$. A conventional merging strategy typically seeks a solution within the $1D$ linear span of these vectors:
\begin{equation}
\tau_{mrg}(\alpha) = \alpha \cdot \tau_t + (1 - \alpha) \cdot \tau_{t-1},
\end{equation}
where $\alpha \in [0, 1]$. In unconstrained merging paradigms, knowledge integration typically occurs within a $t$-dimensional subspace spanned by the full history of task vectors $\{\tau_1, \dots, \tau_t\}$. This high-dimensional subspace provides sufficient degrees of freedom to mitigate inter-task interference and identify optimal consensus points. Under the strict constraints of CL, however, the admissible search subspace undergoes a catastrophic dimensionality collapse, shrinking to a one-dimensional linear trajectory between $\tau_{t-1}$ and $\tau_t$. In this situation, overcoming the three core challenges we identified in our analysis becomes extremely difficult.

To alleviate this limitation, we introduce a perturbation vector $P$ that is orthogonal to $\mathrm{span}(v_{t-1}, v_t)$ to provide essential optimization slack,\
\begin{equation}
P = \text{Normalize} \left( \tilde { r } -  \frac { \langle \tilde { r } , d \rangle } { \| d \| ^ { 2 } } d\right) , \quad \tilde { r } \sim { \cal N } ( 0 , I ),
\end{equation}
where $d  = \tau_{t}-\tau_{t-1}$. This controlled expansion of dimensionality is not intended to induce random fluctuations, but rather to compensate for subspace information lost due to storage constraints. By introducing additional lateral degrees of freedom, the TRM framework is able to deviate from a rigid linear trajectory and explore more elastic regions of the non-convex parameter subspace, which both preserve existing stability and restore responsiveness to future tasks. The resulting augmented merged task vector $\tau_{mrg}$ is formulated as
\begin{equation}
\tau^{mrg}_t(\alpha,\beta)  = \alpha \cdot \tau _ { t } + ( 1 - \alpha ) \cdot \tau _ { t-1 } + \beta \cdot P, 
\end{equation}
where $\alpha, \beta \in \mathbb{R}$ are the interpolation and expansion coefficients, respectively. Finally, the merged model parameters for task $t$ are obtained by updating the initial PTM,
\begin{equation}
\theta_t := merge(\theta_{t-1},\widetilde{\theta_t}) = \theta_{init} +  \tau^{mrg}_t.
\end{equation}

To identify the optimal coefficients $\{\alpha, \beta\}$ within our augmented subspace, we formulate a multi-objective supervisory function. This objective is composed of three synergistic constraints.
\subsection{Task Alignment}
As identified in our analysis, vanilla merging methods often displace the model from the high precision basins of the current task. To counteract this erosion of local optimality, we explicitly enforce task alignment by minimizing the empirical risk on the current task. This objective ensures that the trajectory search remains anchored to the the latest expertise,
\begin{equation}
\mathcal{L}_{align} = \mathbb{E}_{{(x,y) \sim \mathcal{D}^t}} \left[ \mathcal{L}{ce} \big( f(x; \theta_{t-1,t}^{mrg}), y \big) \right].
\label{eq:l_align}
\end{equation}
\subsection{Prediction Consistency}
To mitigate the structural semantic drift, we introduce a prediction consistency objective. This functional regularizer ensures that the merged model's latent representations do not deviate from the collective expertise of its constituent components. Specifically, we define this consistency as the layer-wise discrepancy between the merged model and the functional centroid formed by the current finetuned $\widetilde{\theta_t}$ and the previous consolidated model $\theta_{t-1}$ ,
\begin{equation}
\begin{aligned}
\mathcal{L}_{pre} = \mathbb{E}_{x \sim \mathcal{D}^t} \bigg[& \sum_{l=1}^{L} \omega_l  \Big\|  h^l(x; \theta^{mrg}_{t-1,t}) \\
& - \frac{1}{2} \big( h^l(x; \widetilde{\theta_t}) + h^l(x; \theta_{t-1}) \big) \Big\|_2^2 \bigg],
\end{aligned}
\end{equation}
where $h^l(x; \theta)$ denotes the hidden representations of the $l$-th layer. To account for the architectural characteristic that task-specific semantics are predominantly captured by higher layers \cite{DBLP:conf/iclr/ZhengCQ025}, we employ a progressive layer-wise weighting scheme, 
\begin{equation}
\omega_l = \frac{\exp\left(\max\{1, l-7\}\right)}{\sum_{i=1}^{L} \exp\left(\max\{1, i-7\}\right)}. 
\end{equation}
The choice of the 7th layer as the dividing point is based on the results shown in Figure 2.
\subsection{Gradient Responsiveness}
To re-activate the model's adaptive capacity, we propose the gradient responsiveness objective. We aim to steer the merged model away from pathological dead zones and towards regions with robust and stable update signals. Consider the first-order Taylor expansion of the loss function during a single gradient descent step, we obtain the classical approximation\cite{boyd2004convex,nocedal2006numerical,goodfellow2016deep}:
\begin{equation}
\begin{aligned}
\mathcal{L}_{ce}(\theta^{+}) \approx & \mathcal{L}_{ce}(\theta)  
- \underbrace{\eta\|\nabla_{\theta}\mathcal{L}_{ce}(\theta)\|_{2}^{2}}_{first-order} \\
+ &
\underbrace{\frac{\eta^{2}}{2} {\nabla_{\theta}\mathcal{L}_{ce}(\theta)}^{\top} H(\hat{\theta}) {\nabla_{\theta}\mathcal{L}_{ce}(\theta)}}_{second-order}
\end{aligned}
\end{equation}
where $H ( \hat { \theta } )$ is the Hessian matrix evaluated at some point $\hat { \theta }$ on the line segment connecting $\theta$ and $\theta^{+}$, $\eta$ is the learning rate. This relationship demonstrates that, for a sufficiently small $\eta$, the potential for loss reduction is directly proportional to the squared gradient norm $\|\nabla_{\theta}\mathcal{L}_{ce}(\theta)\|_{2}^{2}$. Consequently, a high gradient norm indicates that the model resides in a highly-responsive manifold characterized by superior learnability and rapid adaptability to new task data. Conversely, a vanishing gradient norm implies that the model is trapped in a pathologically flat region or a dead zone where the optimizer loses its navigation signal. By explicitly maximizing this norm, we ensure that the merged model maintains the kinetic responsiveness required for rapid adaptation to subsequent tasks. The responsiveness regularizer is thus formulated as
\begin{equation}
\mathcal{L}_{res} = -\|\nabla_{\theta} \mathcal{L}_{ce}(\theta_{t-1,t}^{mrg})\|_{2}^{2}
\end{equation}
It's important to emphasize that, under strict CL constraints, data $\mathcal{D}^{t+1}$ from future task is unavailable. However, in standard CL benchmarks, e.g., ImageNet-R, subsequent tasks often share substantial overlap in their low-level feature spaces such as edges, textures, and structural gradients. Mathematically, if $\mathcal{D}^t$ and $\mathcal{D}^{t+1}$ have substantial overlapping regions in low-level feature space, the model responses to different tasks should exhibit the same trend. We believe that the local landscape geometry on $\mathcal{D}^t$ serves as a reliable proxy for the model's optimization kinetics on $\mathcal{D}^{t+1}$. By re-activating responsiveness on the current task, we effectively preserve the model's plasticity for future knowledge acquisition.

Finally, we consolidate the individual constraints into a unified Trajectory Regularization objective. This joint function serves as the supervisory signal to identify the optimal coefficients $\{\alpha, \beta\}$ within our augmented subspace:
\begin{equation} 
\mathcal{L}_{total} = \mathcal{L}_{align} + \lambda_1 \mathcal{L}_{pre} + \lambda_2 \mathcal{L}_{res}, \label{eq:l_total} 
\end{equation}

\begin{table*}[t]
\centering
\caption{Comparison experiments on different benchmarks, \textbf{bolded} indicates optimal, \underline{underlined} indicates sub-optimal.}
\renewcommand{\arraystretch}{1}
\begin{tabular}{l|l|ccc|ccc|ccc}
\toprule
 & \multirow{2}{*}{\textbf{Method (Year)}} &  \multicolumn{3}{c|}{\textbf{CIFAR100}} & \multicolumn{3}{c|}{\textbf{ImageNet-R}} & \multicolumn{3}{c}{\textbf{Stanford-Cars}} \\
 &  & 5 & 10 & 20 & 5 & 10 & 20 & 5 & 10 & 20 \\
\midrule

    & Zero-shot      & \multicolumn{3}{c|}{66.7} &\multicolumn{3}{c|}{72.2} & \multicolumn{3}{c}{64.4} \\
    & Joint          & \multicolumn{3}{c|}{89.4} & \multicolumn{3}{c|}{86.5} & \multicolumn{3}{c}{84.4} \\
\midrule
\multirow{2}{*}{Conventional}
    % & Finetune       & 15.7 & 10.3 & 5.6  & 16.5 & 9.9  & 6.2  & 15.4 & 10.2 & 6.8  \\
    & LwF (2017)         & 68.2 & 67.4 & 67.0 & 66.3 & 60.9 & 55.6 & 65.2 & 61.7 & 57.5 \\
    & EWC (2017)          & 69.3 & 64.8 & 57.3 & 74.5 & 66.7 & 54.1 & 65.8 & 61.3 & 58.0 \\
\midrule
\multirow{5}{*}{PEFT}
    & L2P (2022)           & 77.2 & 73.1 & 68.5 & 77.9 & 75.7 & 72.5 & 67.8   & 66.7   & 65.0   \\
    & DualPrompt (2023)     & 77.4 & 73.6 & 70.4 & 78.6 & 75.9 & 73.4 & 68.9  & 67.1   & 65.2   \\
    & CODAPrompt (2023)     & 77.3 & 75.0 & 74.1 & 78.9 & 76.0 & 73.7 & 67.4   &  67.4  & 64.9   \\
    & RAPF (2024)         & 79.2 & \underline{79.0} & \textbf{79.2} & 80.1 & 79.5 & 78.3 & --   & --   & --   \\
    & CLG-CBM (2025)        & 78.1 & 76.8 & 75.9 & 79.2 & 78.8 & 77.5& \underline{72.8}   & \underline{68.7}   & 64.2    \\
\midrule
\multirow{4}{*}{Model Merging}
    & Model Stock (2024)    & 76.5  & 70.7   & 68.3   & 79.6   & 77.1  & 76.5   & 66.6   & 64.5   & 64.1   \\
    & TIES (2023)           & 79.8   & 76.3   & 73.6   & 81.5   & 79.5   & 78.4   & 67.9   & 66.1   & 65.3   \\
    & MagMax (2024)         & {80.4}   & 75.8  & 72.9   & \underline{82.2}   & \underline{80.1}   & \underline{79.3}   & 68.5   & 65.9   & \underline{65.7}   \\
    & PM (2025)            & --   & --   & --   & 81.4   & 79.9   & 76.3    & --   & --   & --
       \\ 
    &BECAME (2025)          & \underline{80.9}   & 75.2   &  73.5   & 82.1   & 79.8   & 78.7   & --   & --   & --   \\
\midrule
\rowcolor{gray!15}
     & \textbf{TRM (Ours)   }        & \textbf{83.5} & \textbf{80.5} & \underline{78.6} & \textbf{83.6} & \textbf{83.2} & \textbf{82.7} & \textbf{73.2} & \textbf{70.4} & \textbf{66.9} \\
\bottomrule
\end{tabular}
\label{tab1}
\end{table*}

In the actual merging phase, to prevent early convergence to a local optimum during optimization, we initialize $\theta_{init}$ in Equation (6) with a randomly generated point in the search space by randomly place a part parameters in $\theta_{0}$ from the corresponding positions in $\theta_{t-1}$ and $\tilde{\theta_{t}}$.

\section{Experiments}
\subsection{Experiments Setttings}
\subsubsection{Datasets.} Following \cite{marczak2024magmax}, we selected three widely used datasets, CIFAR100 \cite{krizhevsky2009learning}, ImageNet-R \cite{hendrycks2021many} and fine-grained Stanford Cars \cite{krause20133d} for class incremental learning (CIL), and we divided each dataset into 5, 10, and 20 tasks. 

\subsubsection{Metrics.} We use the standard metrics in the continual learning methods to measure performance, last accuracy, which calculates all seen classes' accuracy after the final task and average forgetting, which measures the average drop in accuracy for each task from its peak performance to its state after the final task.

\subsubsection{Comparison Methods.} We compared our method against current state-of-the-art methods, including traditional method, LwF \cite{li2017learning} and EWC \cite{kirkpatrick2017overcoming}, as well as methods based on PEFT, L2P \cite{wang2022learning}, DualPrompt \cite{wang2022dualprompt}, CODAPrompt \cite{smith2023coda}, RAPF \cite{huang2024class} and CLG-CBM \cite{yu2025language}. We also include traditional model merging methods, Model Stock \cite{jang2024model} and  \cite{yadav2023ties}, and the model merged methods specifically designed for CL, MagMax \cite{marczak2024magmax}, PM \cite{qiu2025train} and BECAME \cite{li2025became}. Except for BECAME, which can only operate when the Fisher information matrix is retained, all other methods strictly adhere to the CL constraints mentioned in this paper. In the experiments, the backbones of all compared methods are kept consistent with ours, and all results are obtained by executing in the same environment and strictly adhering to the source code parameters.

\subsubsection{Implementation Details.} The image encoder is CLIP of ViT-B/16 from OpenAI, and the training batch size is 128, the learning rate is $1 \times 10^{-5}$, and cosine annealing learning rate schedule and AdamW optimizer with weight decay 0.1. The training epoch is $20$ and merging epoch is $5$ for all dataset. We set $\lambda_1 = 0.1$, $\lambda_2 = 0.01$. We use the CLIP's text encoder to encode labels and use its output as a classifier. Except for the image encoder, all other components are kept frozen throughout training.   

\subsection{Experimental Results}
Experimental results for CIL are shown in Table \ref{tab1} (average forgetting are shown in the supplementary materials). On the CIFAR100 dataset, our method achieves accuracy of $83.5\%$ and $80.5\%$ in the 5 and 10 tasks respectively, representing improvements of $3.1\%$ and $1.5\%$ over the previous SOTA method. After learning 20 consecutive tasks, the performance is slightly lower than the optimal RAPF. On the ImageNet-R, our method achieves the best performance across all settings, with accuracies of $83.6\%$, $83.2\%$, and $82.7\%$ for 5, 10, and 20 tasks, corresponding to improvements of $1.4\%$, $3.1\%$, and $3.4\%$ over previous SOTA methods. On the fine-grained Stanford Cars dataset, our method achieves $73.2\%$, $70.4\%$, and $66.9\%$ for all settings, representing gains of $0.4\%$, $1.7\%$, and $1.2\%$, respectively. In summary, the proposed method effectively mitigates catastrophic forgetting under continuous data streams without storing any previous model or data distribution.
\begin{table}[t]
\centering
\caption{{Ablation Study on ImageNet-R (10 tasks).} (a) is the stochastic parameter crossover baseline. We evaluate Task Alignment ($\mathcal{L}_{task}$), Prediction Consistency ($\mathcal{L}_{pre}$), and Gradient Responsiveness ($\mathcal{L}_{res}$).}
\vspace{1mm}
\footnotesize % 调小字体
\setlength{\tabcolsep}{5pt} % 进一步收缩列间距
\renewcommand{\arraystretch}{1.05} % 稍微控制行高
\begin{tabular}{l c c c c}
\toprule
\textbf{Variant} & \textbf{$\mathcal{L}_{task}$} & \textbf{$\mathcal{L}_{pre}$} & \textbf{$\mathcal{L}_{res}$} & \textbf{Acc (\%)} \\
\midrule
(a) Baseline & -- & -- & -- & 79.3 \\
\midrule
(b) & \checkmark & & & 71.8 \\
(c) & & \checkmark & & 72.4 \\
(d) & & & \checkmark & 76.5 \\
\cmidrule(lr){1-5}
(e) & \checkmark & \checkmark & & 70.7 \\
(f) & \checkmark & & \checkmark & 81.9 \\
(g) & & \checkmark & \checkmark & 80.6 \\
\midrule
\rowcolor{gray!15}
(h) \textbf{TRM} & \checkmark & \checkmark & \checkmark & \textbf{83.2} \\
\bottomrule
\end{tabular}
\label{tab3}
\end{table}
\subsection{Ablation Study}
To evaluate the contribution of each component within the TRM framework, we conduct ablation study on the ImageNet-R with 10 tasks. As shown in Table \ref{tab3}, baseline randomly replaced the parameters of ${\theta_{init}}$ with those at the corresponding positions of $\theta_{t-1}$ and $\tilde\theta_{t}$ achieves only $79.3\%$. Within the TRM framework, task alignment and prediction consistency are designed to safeguard the model's stability, while gradient responsiveness  is dedicated to restoring its plasticity. When these constraints are applied in isolation, they induce an unbalanced optimization bias that skews the search process toward suboptimal regions of subspace, resulting in performance decreases of $7.5\%$, $6.9\%$, and $2.8\%$, respectively. When the two stability constraints are used together, performance drops by $8.6\%$. When we fix the gradient responsiveness constraint and pair it with either the task alignment or prediction consistency, performance improves by $2.6\%$ and $1.3\%$ compared to random merged. When all three constraints are used simultaneously, performance reaches $83.2\%$.

\begin{figure}[h]
    \centering
    \includegraphics[width=1\columnwidth]{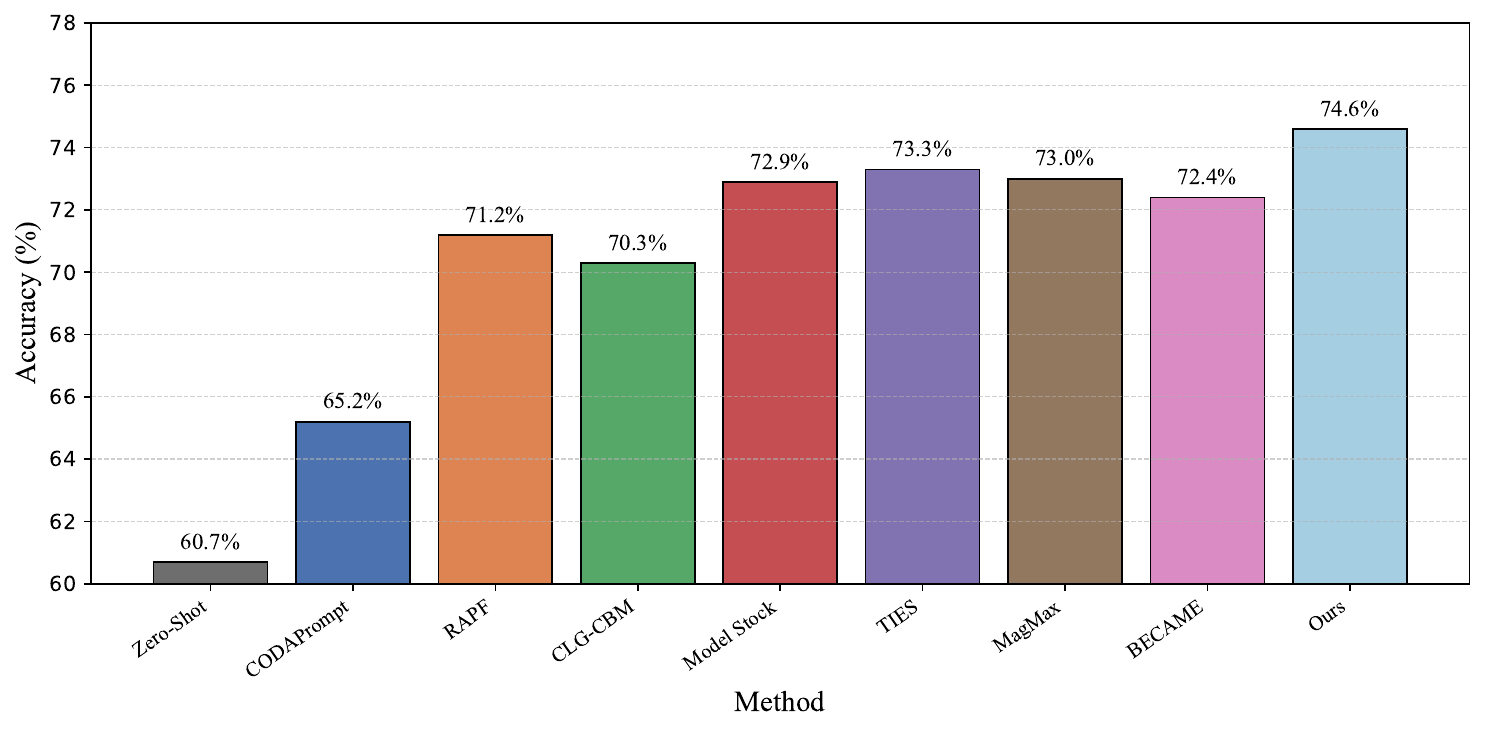}
    \caption{Experiments on ImageNet1K are conducted by evenly dividing the 1000 classes into 100 distinct training tasks.}
    \label{fig6}
\end{figure}

\subsection{Further Analysis}
\subsubsection{Large Scale Dataset.}
We conducted experiments on the large-scale ImageNet1K dataset by evenly dividing its 1000 classes into 100 non-overlapping tasks. The experimental results are presented in Figure \ref{fig6}. As the number of tasks increases, the difficulty of learning on continue data stream grows significantly; nevertheless, our method consistently achieves the best performance, demonstrating its scalability and robustness to data scale.
\begin{figure}[t]
    \centering
    \includegraphics[width=1\columnwidth]{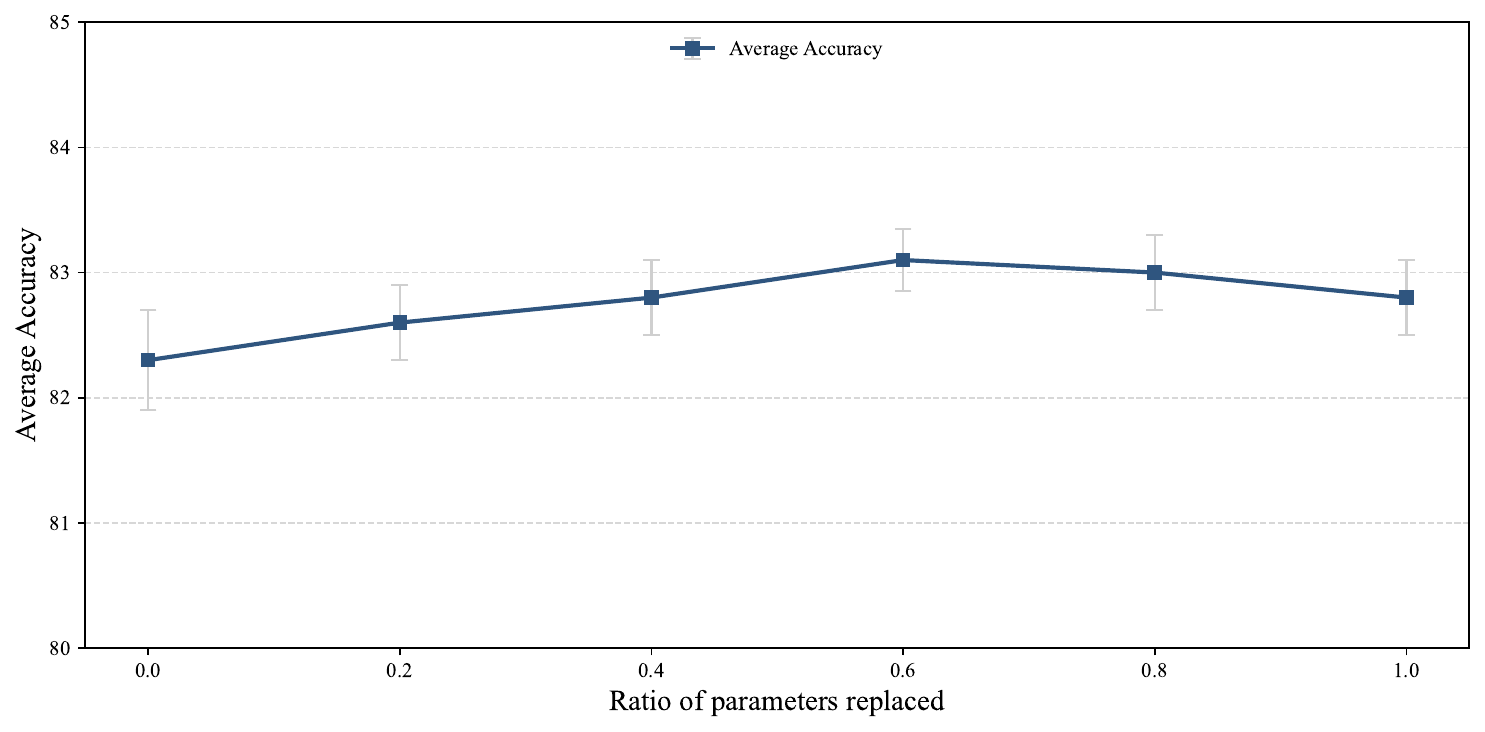}
    \caption{Ratio of replaced parameters. The experiments were conducted on ImageNet-R with 10 tasks, and for each ratio, we conducted 10 independent training runs and recorded the maximum, minimum, and average performance.}
    \label{fig7}
\end{figure}
\subsubsection{The Ration of Replaced Parameters.} 
At the end of the methods section, we stated that during training, we randomly replaced a portion of the parameters in the initial model $\theta_{init}$ with parameters from the corresponding positions in models $\theta_{t-1}$ and $\tilde\theta_{t}$ to ensure optimization stability. To validate this, we conducted experiments on 10 tasks from the ImageNet-R dataset. In experiments, the position of the replaced parameters and the source of the replacement (from $\theta_{t-1}$ or $\tilde\theta_{t}$) are completely random, only the overall replacement ratio was controlled. For each ratio, we conducted 10 independent training runs and recorded the maximum, minimum, and average performance. The experiments results are shown in the Figure \ref{fig7}. As observed, using the original parameters as the initialization led to slightly lower performance and exhibited a wider range between the best and worst results compared to experiments with partial parameter replacement. As the replacement ratio increased, the variance in performance initially decreased and then gradually increased. Based on these findings, we set the parameter replacement ratio to $0.6$ in our experiments.
\begin{table}[h]
\centering
\caption{Analysis of optimization stability. $\lambda_{\max}$ represents the Hessian spectral norm on ImageNet-R.}
\label{tab:lambda_sensitivity}
% \small % 降为 footnotesize，比 small 更小
\setlength{\tabcolsep}{6pt} % 紧凑的列间距
\renewcommand{\arraystretch}{1.0} % 恢复标准行高
\begin{tabular}{l c}
\toprule
\textbf{Configuration} & Hessian Spectral Norm ($\lambda_{\max}$) \\ 
\midrule
Finetune  & 8.60 \\ 
\midrule
$\lambda_2 = 0$  & 1.15 \\
\rowcolor{gray!15}
\textbf{$\lambda_2 = 0.01$ (Ours)} & \textbf{5.85} \\
$\lambda_2 = 1.0$ & 45.20 \\
$\lambda_2 = 10.0$ &  96.50 \\ 
\bottomrule
\end{tabular}
\end{table}
\subsubsection{Analysis Of Optimization Stability.} 
To assess the stability of the merged model, we quantify the local landscape curvature by computing the max eigenvalue $\lambda_{\max}$ of the Hessian matrix. In optimization theory \cite{keskar2016large}, $\lambda_{\max}$ serves as a direct proxy for the sharpness of the loss surface, where excessively high values often correlate with poor generalization and numerical instability. We conduct experiments on the ImageNet-R, 10 tasks with different responsiveness coefficient $\lambda_2$, $\lambda_2 = 0.01$ identified as the optimal balance and Finetune indicates $\tilde{\theta_t}$ without merged. Our empirical results in Tabel \ref{tab:lambda_sensitivity} indicates that while TRM increases the gradient norm to provide optimization impetus, it maintains $\lambda_{\max}$ within a moderate and stable range and does not induce pathological sharpness.
\begin{table}[t]
\centering
\caption{Ablation on Search Space Dimensionality. We evaluate on ImageNet-R with 10 tasks}
\label{tab:dim_ablation}
% \small
\setlength{\tabcolsep}{10pt}
\renewcommand{\arraystretch}{1.0}
\begin{tabular}{l c c}
\toprule
\textbf{Subspace Components} & \textbf{Acc (\%)} & \textbf{Time (min)} \\ 
\midrule
$\tau_{t-1}, \tau_{t}$ & 79.8 & 25 \\
\rowcolor{gray!15}
$\tau_{t-1}, \tau_{t}, P$ (Ours) & \textbf{83.2} & \textbf{33} \\
$\tau_{t-1}, \tau_{t}, P, P_1$ & 83.4 & 42 \\
$\tau_{t-1}, \tau_{t}, P, P_1, P_2$ & 83.5 & 58 \\ 
\bottomrule
\end{tabular}
\end{table}
\subsubsection{Search Space Dimensionality}
To investigate how the degrees of freedom within the augmented trajectory subspace influence merging performance, we conducted an ablation study by scaling the subspace dimensionality. For expansion into higher dimensions, we introduce additional perturbation vectors $\{P_1, {P}_2, \dots \}$. To ensure that each new dimension provides maximal optimization slack without interfering with the established task representations, we enforce mutual orthogonality across all basis vectors. Specifically, we employ a Gram-Schmidt process to generate each $P_i$ such that $P_i \perp span(\tau_t, \tau_{t-1}) \quad \text{and} \quad P_i \perp P_j, \quad \forall j < i$. The experiments were conducted on ImageNet-R with 10 tasks. As summarized in Table \ref{tab:dim_ablation}, during the process of increasing dimensions, the performance did not improve as expected, despite a significant increase in training time. Consequently, our augmented trajectory subspace only introduced a single perturbation direction $P$.
\begin{table}[h]
\centering
\caption{Performance on ImageNet-R with different numbers of tasks using ViT-B/16 pre-trained on LAION-400M.}
\begin{tabular}{lccc}
\toprule
\multirow{2}{*}{Method} & \multicolumn{3}{c}{ImageNet-R} \\
\cmidrule(lr){2-4}
& 5 tasks & 10 tasks & 20 tasks \\
\midrule
Model Stock & 79.6 & 77.0 & 76.4 \\
TIES        & 81.7 & 79.3 & 78.4 \\
MagMAX      & 82.2 & 80.4 & 79.2 \\
BECAME      & 81.2 & 80.0 & 76.1 \\
\rowcolor{gray!15}
TRM (Ours)        & 83.4 & 83.1 & 82.5 \\
\bottomrule
\end{tabular}
\label{tab000}
\end{table}
\subsubsection{Different Backbone}
In the main experiment Tabel \ref{tab1}, we adopted ViT-B/16 as the backbone. To evaluate the applicability of our method across different visual encoders, we further introduced two alternative backbones for testing, ViT-L/14 (pre-trained on WebImageText) and ViT-B/16 (pre-trained on LAION-400M). All experiments were conducted on the ImageNet-R with different tasks. As shown in Table \ref{tab000} and Table \ref{tab0000}, our method consistently achieves the best performance across different backbones, demonstrating the robustness and versatility of our proposed method.
\begin{table}[t]
\centering
\caption{Performance on ImageNet-R with different numbers of tasks using ViT-L/14 pre-trained on WebImageText.}
\begin{tabular}{lccc}
\toprule
\multirow{2}{*}{Method} & \multicolumn{3}{c}{ImageNet-R} \\
\cmidrule(lr){2-4}
& 5 tasks & 10 tasks & 20 tasks \\
\midrule
Model Stock & 87.2 & 86.8 & 86.7 \\
TIES        & 87.8 & 87.4 & 86.9 \\
MagMAX      & 89.7 & 89.4 & 89.1 \\
BECAME      & 87.9 & 87.8 & 87.1 \\
\rowcolor{gray!15}
TRM (Ours)        & 90.8 & 90.5 & 89.9 \\
\bottomrule
\end{tabular}
\label{tab0000}
\end{table}
\subsubsection{Time Consumption.}
The proposed TRM may introduce additional training time, therefore, we compared the runtime across ImageNet-R and CIFAR100 with 10 tasks. The experimens about running time (minutes) are shown in Table \ref{tab4}. According to the results, our method significantly improves performance while introducing minimal training time.
\begin{table}[h]
\centering
\caption{Training time across different datasets.}
\resizebox{\linewidth}{!}{
\begin{tabular}{l|ccccc}
\toprule
 & Model Stock & TIES & MagMax & BECAME & Ours \\
\midrule
ImageNet-R & 30 & 30 & 29 & 87 & 33 \\
CIFAR100   & 42 & 42 & 42 & 92 & 44 \\
\bottomrule
\end{tabular}
}
\label{tab4}
\end{table}
\section{Conclusion}
In this paper, we investigate the limitations of existing model merging methods in realistic continual learning scenarios, where only the pre-task and post-task models are accessible, and the merged model must serve as the initialization for future training. Our analysis reveals that existing model merging method often prioritize global optimization, while neglecting local task-specific adaptation, resulting in accumulated errors and slow gradient evolution during training.To address these challenges, we propose a trajectory regularized merging framework, reformulating model merging as an optimal point search within the subspace spanned by the task vectors of the current and previous tasks, guided by three complementary objectives, task alignment, prediction consistency, and gradient responsiveness, to improve the stability and plasticity of the merged model. Extensive experiments across multiple benchmarks confirm the effectiveness of our method under strict continual learning constraints, achieving state-of-the-art performance.

\bibliography{main}
\bibliographystyle{icml2026}

%%%%%%%%%%%%%%%%%%%%%%%%%%%%%%%%%%%%%%%%%%%%%%%%%%%%%%%%%%%%%%%%%%%%%%%%%%%%%%%
%%%%%%%%%%%%%%%%%%%%%%%%%%%%%%%%%%%%%%%%%%%%%%%%%%%%%%%%%%%%%%%%%%%%%%%%%%%%%%%
% APPENDIX
%%%%%%%%%%%%%%%%%%%%%%%%%%%%%%%%%%%%%%%%%%%%%%%%%%%%%%%%%%%%%%%%%%%%%%%%%%%%%%%
%%%%%%%%%%%%%%%%%%%%%%%%%%%%%%%%%%%%%%%%%%%%%%%%%%%%%%%%%%%%%%%%%%%%%%%%%%%%%%%
\newpage
\appendix
\onecolumn
\section{Supplement to Introduction}
\begin{table*}[h]
\centering
\caption{Comparison of different merging methods on ImageNet-R with 10 and 20 tasks.}
\begin{tabular}{c|cccc|cccc}
\toprule
\multirow{2}{*}{Implementation} & \multicolumn{4}{c|}{ImageNet-R (10 Tasks)} & \multicolumn{4}{c}{ImageNet-R (20 Tasks)} \\
\cmidrule(lr){2-5} \cmidrule(lr){6-9}
& TIES & Model Stock & MagMax & BECAME & TIES & Model Stock & MagMax & BECAME\\
\midrule
A & 82.9 & 83.3 & 83.3 & - & 82.2 & 82.1 & 82.2 & - \\
B & 80.6 & 81.8 & 83.3 & 79.8 & 81.3 & 80.7 & 82.2 & 78.7 \\
C & 79.5 & 77.1 & 80.1 & - & 78.4 & 76.5 & 79.3 & - \\
 \bottomrule
\end{tabular}
\label{stab1}
\end{table*} 
We evaluated the impact of storage constraints on existing model merging methods. Specifically, experiments were conducted on the ImageNet-R dataset with 10 and 20 tasks. Implementation A,B,C indicate storing all past models, storing only feature vectors or distribution statistics, and strictly constrained storage—corresponding respectively. Note that BECAME can only operate when the Fisher information matrix is retained and MagMax performs merging by selecting the maximum parameter value at each position, which should yield consistent results regardless of whether merging is performed in a single step using all previous task vectors or incrementally at each stage. The results, presented in Table \ref{stab1}, show a clear trend, the performance of existing merging methods in continuous data-stream scenarios is strongly correlated with the amount of accessible task-specific information. As available memory decreases, overall performance consistently deteriorates.
\section{Results of Figure 1-3}
In the analysis section of the manuscript, we conducted three distinct experiments to further investigate the plasticity and stability of the merged model. The corresponding results are presented in Figures 1-3, all of which include our proposed method for comparative analysis. Below, we provide a detailed interpretation of our method’s performance. 

Figure 1 illustrates the loss landscape during finetuning and approximates the projection of the merged model based on the loss magnitude on the current task. It is evident that our proposed TRM effectively reduces task-specific loss, although it does not converge precisely to the lowest loss region. Figure 2 visualizes the consistency of outputs across layers in the image encoder for the same input, measured by the L2 norm. The results demonstrate that our method significantly reduces inter layer output discrepancies, indicating greater continuity in the parameter space, thus improve the model stability. Figure 3 presents the early stage training performance of each merging method on subsequent tasks. Compared to other methods, our method shows faster adaptation during the initial training phases, reflecting stronger plasticity at initialization.
\section{More Experiments}
\subsection{Average Forgetting}
In addition to the final accuracy shown in Table 1, we also calculated the average forgetting. The experimental results are shown in the Table \ref{tab_forgetting}, and it can be seen that our method still performs the best.
\subsection{Domain Incremental Learning.}
In Section 4.3, we established gradient responsiveness objective relies on the assumption of overlap-
ping regions in low-level feature space between successive tasks. To rigorously evaluate the limits of this proxy hypothesis under substantial distribution shifts, we extend our evaluation to Domain Incremental Learning (DIL) benchmarks. Unlike standard CIL, DIL presents a more challenging landscape where the model must adapt to significant domain-level variations while preserving categorical knowledge. We selected ImageNet-R and DomainNet \cite{peng2019moment} for domain incremental learning (DIL). DomainNet is divided into $6$ independent tasks based on domain categories, while ImageNet-R is split into $15$ independent tasks. Similarly, we split these datasets into the corresponding number of tasks following the setting of CIL for a fair comparison of CIL and DIL performance.

In Table \ref{tab2}, we report results only on the complete test set for both CIL and DIL settings, without evaluating performance on individual domains. As shown, our method consistently outperforms all comparison approaches across all scenarios. On DomainNet, our method achieves accuracies of 69.5\% and 70.3\% under the CIL and DIL settings, respectively. On ImageNet-R, the corresponding results are $83.1\%$ and $84.9\%$. These results confirm that the Trajectory Regularization objective acts as a scenario-agnostic catalyst, ensuring that the model maintains its plasticity and stability across different training scenarios.
\begin{table*}[t]
\centering
\caption{Comparison of Average Forgetting on different benchmarks. Lower values indicate better knowledge retention. \textbf{Bolded} indicates optimal, \underline{underlined} indicates sub-optimal.}
\label{tab_forgetting}
% \small
% \setlength{\tabcolsep}{6pt}
\renewcommand{\arraystretch}{1.2} % 稍微增加行高，避免数字拥挤
\begin{tabular}{l| l| ccc |ccc |ccc} % 移除内部竖线，符合 ICML/CVPR 审美
\toprule
& \multirow{2}{*}{\textbf{Method (Year)}} & \multicolumn{3}{c}{\textbf{CIFAR100}} & \multicolumn{3}{c}{\textbf{ImageNet-R}} & \multicolumn{3}{c}{\textbf{Stanford-Cars}} \\
\cmidrule(lr){3-5} \cmidrule(lr){6-8} \cmidrule(lr){9-11}
& & 5 & 10 & 20 & 5 & 10 & 20 & 5 & 10 & 20 \\
\midrule
\multirow{2}{*}{Conv.}
& LwF (2017) & 18.2 & 22.4 & 26.5 & 15.3 & 19.8 & 24.1 & 16.2 & 20.5 & 25.4 \\
& EWC (2017) & 14.5 & 19.1 & 28.4 & 10.2 & 16.7 & 22.8 & 12.8 & 18.3 & 23.6 \\
\midrule
\multirow{5}{*}{PEFT}
& L2P (2022) & 4.2 & 5.8 & 7.4 & 3.1 & 4.5 & 5.8 & 5.2 & 6.4 & 7.9 \\
& DualPrompt (2023) & 3.8 & 5.1 & 6.9 & 2.8 & 4.1 & 5.2 & 4.9 & 6.1 & 7.5 \\
& CODAPrompt (2023) & \underline{3.2} & 4.5 & 5.8 & 2.5 & 3.8 & 4.9 & 4.5 & 5.8 & 7.1 \\
& RAPF (2024) & 3.4 & \underline{4.2} & \underline{5.2} & \underline{2.1} & \underline{3.2} & \underline{4.4} & -- & -- & -- \\
& CLG-CBM (2025) & 3.9 & 4.8 & 6.2 & 2.6 & 3.9 & 5.1 & \underline{3.8} & \underline{5.2} & \underline{6.8} \\
\midrule
\multirow{4}{*}{Model Merging}
& Model Stock (2024) & 6.5 & 8.7 & 10.4 & 4.6 & 6.8 & 8.5 & 7.2 & 9.5 & 11.2 \\
& TIES (2023) & 5.8 & 7.5 & 9.2 & 3.9 & 5.4 & 7.1 & 6.5 & 8.1 & 9.8 \\
& MagMax (2024) & 5.2 & 7.1 & 8.8 & 3.4 & 4.9 & 6.5 & 6.1 & 7.8 & 9.2 \\
& PM (2025) & -- & -- & -- & 3.6 & 5.0 & 6.8 & -- & -- & -- \\
& BECAME (2025) & 4.8 & 6.9 & 8.5 & 3.1 & 4.6 & 6.2 & -- & -- & -- \\
\midrule
\rowcolor{gray!15}
& \textbf{TRM (Ours)} & \textbf{2.5} & \textbf{3.6} & \textbf{4.2} & \textbf{1.8} & \textbf{2.6} & \textbf{3.5} & \textbf{2.9} & \textbf{4.6} & \textbf{6.0} \\
\bottomrule
\end{tabular}
\end{table*}
\begin{table}[h]
    \centering
    \caption{Comparison of different methods on DomainNet and ImageNet-R under CIL and DIL settings. 
    Higher is better.}
    \vspace{2mm}
    \setlength{\tabcolsep}{8pt}
    \renewcommand{\arraystretch}{1.15}
    \begin{tabular}{lcccc}
        \toprule
        \multirow{2}{*}{\textbf{Method}} 
        & \multicolumn{2}{c}{\textbf{DomainNet}} 
        & \multicolumn{2}{c}{\textbf{ImageNet-R}} \\
        & \textbf{CIL} & \textbf{DIL} & \textbf{CIL} & \textbf{DIL} \\
        \midrule
        LwF          & 57.3 & 59.5 & 58.2 & 61.7 \\
        EWC          & 57.9 & 59.2 & 63.5 & 65.1 \\
        Model Stock  & 63.9 & 70.2 & 77.0 & 82.6 \\
        TIES         & 66.6 & 66.2 & 78.8 & 83.5 \\
        MagMax       & 68.4 & 68.7 & 80.0 & 83.9 \\
        \midrule
        \rowcolor{gray!15}
        \textbf{Ours} & \textbf{69.5} & \textbf{70.3} & \textbf{83.1} & \textbf{84.9} \\
        \bottomrule
    \end{tabular}
    \label{tab2}
\end{table}
\subsection{Hyperparameter Analysis}
\begin{figure}[h]
    \centering
    \includegraphics[width=0.5\columnwidth]{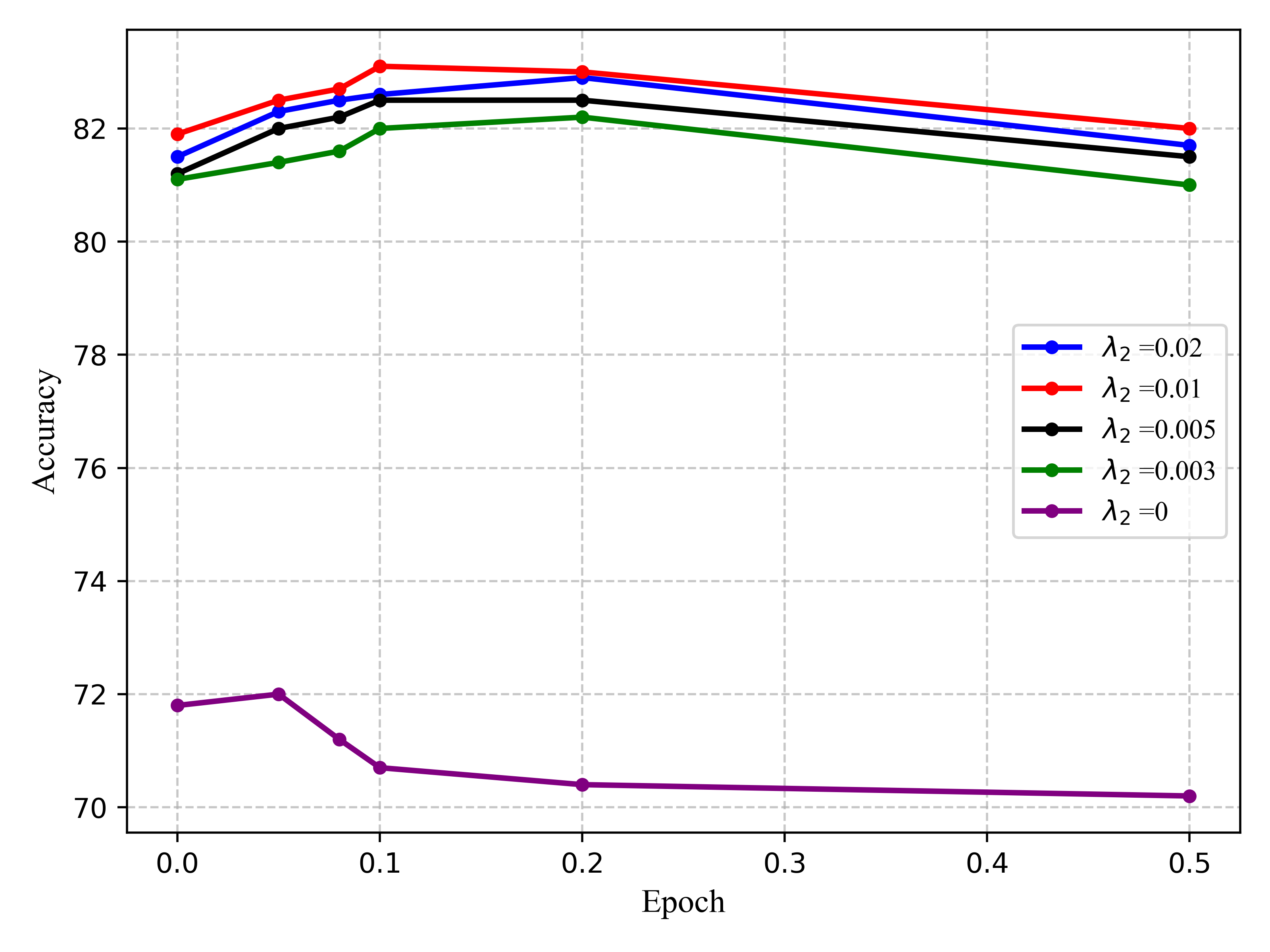}
    \caption{Sensitivity analysis. The horizontal axis represents different values of $\lambda_1$ while the varying line styles indicate different values of $\lambda_2$.}
    \label{sfig1}
\end{figure}

We set $\lambda_1 = 0.1, \lambda_2 = 0.01$ in Equation (12) in our experiments. We conducted sensitivity analysis, all experiments were performed on the ImageNet-R  with 10 tasks. And the experimental results are presented in the Figure \ref{sfig1}. Our experiment results show that the proposed method remains stable across different hyperparameters.
\subsection{Searching Epoch}
\begin{figure}[h]
    \centering
    \includegraphics[width=0.5\columnwidth]{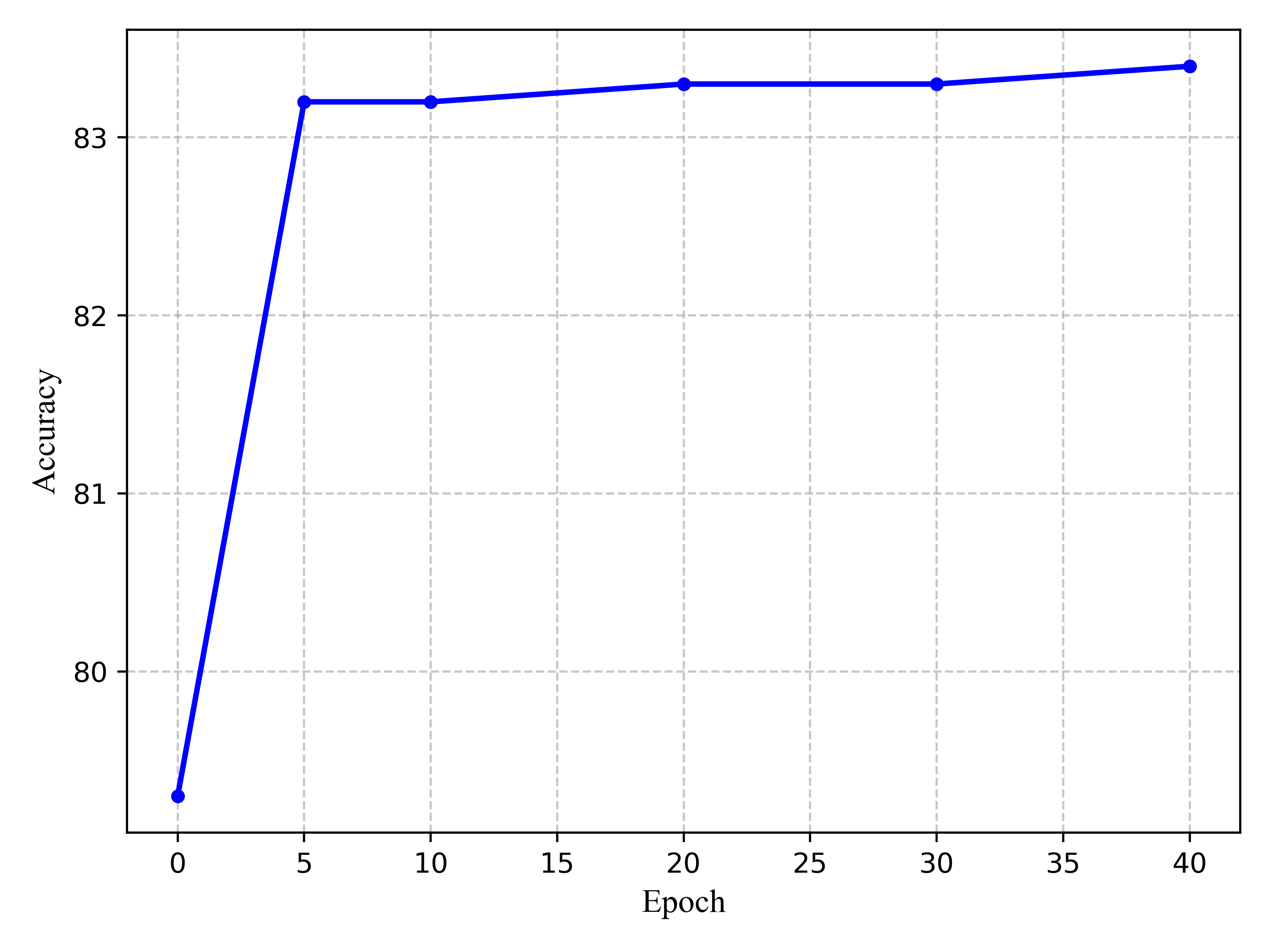}
    \caption{Sensitivity analysis of searching epoch.}
    \label{sfig2}
\end{figure}

We analyze the impact of searching epoch selection, the experiments were conducted on ImageNet-R with 10 tasks and the corresponding experimental results presented in Figure \ref{sfig2}. Because the optimization range of the two parameters $\alpha $ and $\beta$ was kept very small, model performance became stable after $5$ epochs. Therefore, we fixed the epoch is $5$ across all datasets.

%%%%%%%%%%%%%%%%%%%%%%%%%%%%%%%%%%%%%%%%%%%%%%%%%%%%%%%%%%%%%%%%%%%%%%%%%%%%%%%
%%%%%%%%%%%%%%%%%%%%%%%%%%%%%%%%%%%%%%%%%%%%%%%%%%%%%%%%%%%%%%%%%%%%%%%%%%%%%%%

\end{document}